\documentclass[11pt,a4paper]{article}
\usepackage[hyperref]{naaclhlt2018}
\usepackage{times}
\usepackage{latexsym}
\usepackage{amssymb}
\usepackage{amsmath}
\usepackage{url}
\usepackage{breqn}
\usepackage{graphicx}
\usepackage{graphics}

\newcommand{\R}{\mathbb{R}}
\newcommand{\batch}{b}
\newcommand{\heads}{h}

\aclfinalcopy 

\title{Self-Attention with Relative Position Representations}

\author{Peter Shaw \\ Google \\ petershaw@google.com \\
\And
Jakob Uszkoreit \\ Google Brain \\ usz@google.com
\And
Ashish Vaswani \\ Google Brain \\ avaswani@google.com
}

\begin{document}
\maketitle
\begin{abstract}

Relying entirely on an attention mechanism, the Transformer introduced by Vaswani et al.~\shortcite{vaswani2017} achieves state-of-the-art results for machine translation. In contrast to recurrent and convolutional neural networks, it does not explicitly model relative or absolute position information in its structure. 
Instead, it requires adding representations of absolute positions to its inputs.
In this work we present an alternative approach, extending the self-attention mechanism to efficiently consider representations of the relative positions, or distances between sequence elements.
On the WMT 2014 English-to-German and English-to-French translation tasks, this approach yields improvements of 1.3 BLEU and 0.3 BLEU over absolute position representations, respectively. Notably, we observe that combining relative and absolute position representations yields no further improvement in translation quality.
We describe an efficient implementation of our method and cast it as an instance of relation-aware self-attention mechanisms that can generalize to arbitrary graph-labeled inputs.

\end{abstract}

\section{Introduction}

Recent approaches to sequence to sequence learning typically leverage recurrence
~\cite{sutskever2014}, convolution ~\cite{gehring2017, kalchbrenner2016},
attention ~\cite{vaswani2017}, or a combination of recurrence and attention
~\cite{bahdanau2014, cho2014, luong2015, wu2016} as basic building blocks. These approaches incorporate
information about the sequential position of elements differently.

Recurrent neural networks (RNNs) 
typically compute a hidden state $h_t$, as a function of their input at time $t$ and a previous hidden state $h_{t-1}$, capturing relative and absolute positions along the time dimension directly through their sequential structure.
Non-recurrent models do not necessarily consider input elements sequentially and may hence require explicitly encoding position information to be able to use sequence order. 

One common approach is to use position encodings which are combined with input elements to expose position information to the model. These position encodings can be a deterministic function of position ~\cite{sukhbaatar2015, vaswani2017} or learned representations. Convolutional neural networks inherently capture relative positions within the kernel size of each convolution. They have been shown to still benefit from position encodings ~\cite{gehring2017}, however.

For the Transformer, which employs neither convolution nor recurrence, incorporating explicit representations of position information is an especially important consideration since the model is otherwise entirely invariant to sequence ordering. Attention-based models have therefore used position encodings or biased attention weights based on distance ~\cite{parikh2016}.

In this work we present an efficient way of incorporating relative position representations in the self-attention mechanism of the Transformer. Even when entirely replacing its absolute position encodings, we demonstrate significant improvements in translation quality on two machine translation tasks.

Our approach can be cast as a special case of extending the self-attention mechanism of the Transformer to considering arbitrary relations between any two elements of the input, a direction we plan to explore in future work on modeling labeled, directed graphs.

\section{Background}

\subsection{Transformer}

The Transformer ~\cite{vaswani2017} employs an encoder-decoder structure, consisting of stacked encoder and decoder layers.
Encoder layers consist of two sublayers: self-attention followed by a position-wise feed-forward layer.
Decoder layers consist of three sublayers: self-attention followed by encoder-decoder attention, followed by a position-wise feed-forward layer.
It uses residual connections around each of the sublayers, followed by layer normalization ~\cite{ba2016}. The decoder uses masking in its self-attention to prevent a given output position from incorporating information about future output positions during training.

Position encodings based on sinusoids of varying frequency are added to encoder and decoder input elements prior to the first layer.
In contrast to learned, absolute position representations, the authors hypothesized that sinusoidal position encodings would help the model to generalize to sequence lengths unseen during training by allowing it to learn to attend also by relative position. This property is shared by our relative position representations which, in contrast to absolute position representations, are invariant to the total sequence length.

Residual connections help propagate position information to higher layers.

\subsection{Self-Attention}\label{selfattn}

Self-attention sublayers employ $h$ attention heads.
To form the sublayer output, results from each head are concatenated and a parameterized linear transformation is applied.

Each attention head operates on an input sequence, $x = (x_1, \ldots, x_n)$ of
$n$ elements where $x_i \in \R^{d_x}$, and computes a new sequence $z = (z_1, \ldots, z_n)$ of the same length
where $z_i \in \R^{d_z}$.

Each output element, $z_i$, is computed as weighted sum of a linearly transformed input elements:

\begin{equation}\label{eq:attn}
z_i = \sum_{j=1}^{n} \alpha_{ij} (x_jW^V)
\end{equation}

Each weight coefficient, $\alpha_{ij}$, is computed using a softmax function:

\[
\alpha_{ij} = \frac{ \exp{e_{ij}} }{ \sum_{k=1}^{n} \exp{e_{ik}} }
\]

And $e_{ij}$ is computed using a compatibility function that compares two input elements:

\begin{equation}\label{eq:e}
e_{ij} = \frac{(x_iW^Q)(x_jW^K)^T}{\sqrt{d_z}}
\end{equation}

Scaled dot product was chosen for the compatibility function, which enables efficient computation.
Linear transformations of the inputs add sufficient expressive power.

$W^Q$, $W^K$, $W^V \in \R^{d_x \times d_z}$ are parameter
matrices. These parameter matrices are unique per layer and attention head.

\section{Proposed Architecture}

\subsection{Relation-aware Self-Attention}\label{secselfattn}

We propose an extension to self-attention to consider the
pairwise relationships between input elements. In this sense, we model the input as a labeled, directed, fully-connected graph.

The edge between input elements $x_i$ and $x_j$ is represented by vectors $a^V_{ij}, a^K_{ij} \in \R^{d_a}$. The motivation for learning two distinct edge representations is that $a^V_{ij}$ and $a^K_{ij}$ are suitable for use in eq.~\eqref{eq:attn2} and eq.~\eqref{eq:e2}, respectively, without requiring additional linear transformations.
These representations can be shared across attention heads. We use $d_a = d_z$.

We modify eq.~\eqref{eq:attn} to propagate edge information to the sublayer output:

\begin{equation}\label{eq:attn2}
z_i = \sum_{j=1}^{n} \alpha_{ij} (x_jW^V + a^V_{ij})
\end{equation}

This extension is presumably important for tasks where information about the edge types selected by a given attention head is useful to downstream encoder or decoder layers. However, as explored in \ref{secmodelvariations}, this may not be necessary for machine translation.

We also, importantly, modify eq.~\eqref{eq:e} to consider edges when determining compatibility:

\begin{equation}\label{eq:e2}
e_{ij} = \frac{x_iW^Q(x_jW^K+a^K_{ij})^T}{\sqrt{d_z}}
\end{equation}

The primary motivation for using simple addition to incorporate edge representations in eq.~\eqref{eq:attn2} and eq.~\eqref{eq:e2} is to enable an efficient implementation described in \ref{secimpl}. 

\subsection{Relative Position Representations}

\begin{figure}[t!]
\begin{center}
\includegraphics[width=.45\textwidth]{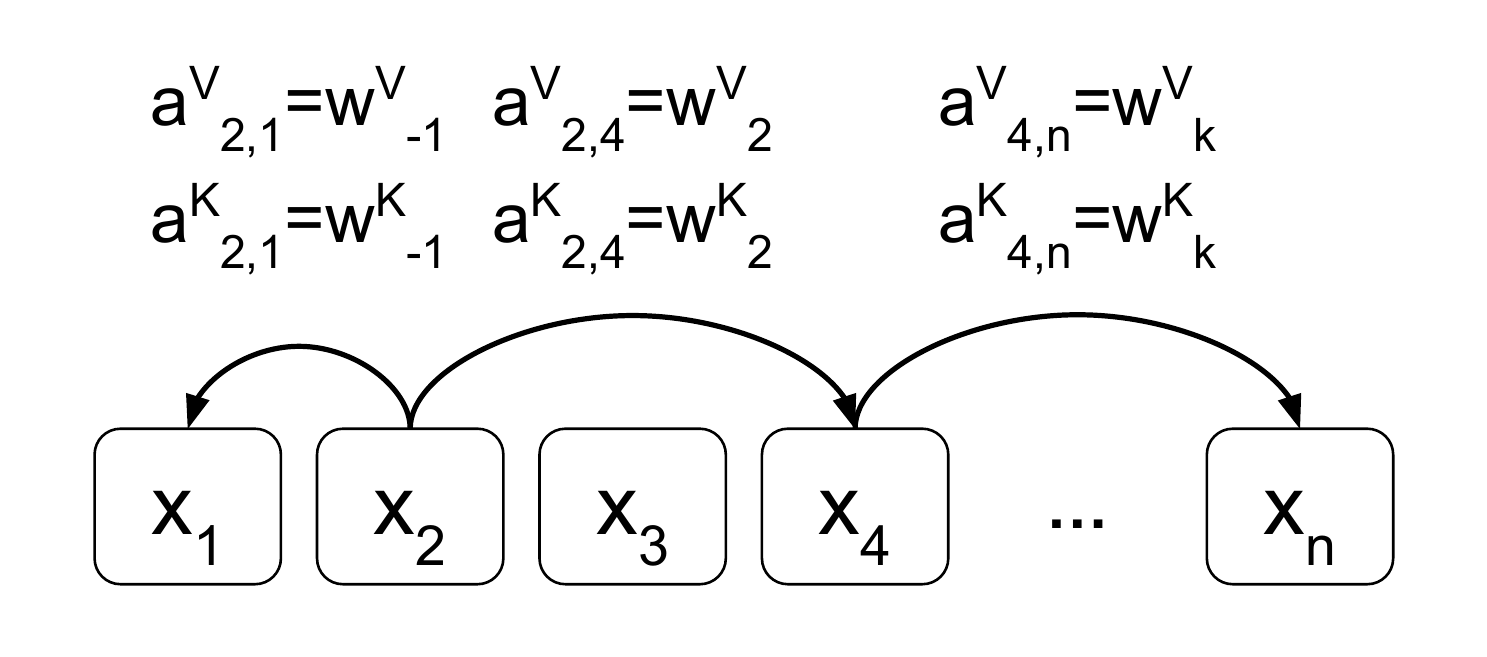}
\caption{Example edges representing relative positions, or the distance between elements. We learn representations for each relative position within a clipping distance $k$. The figure assumes $2 <= k <= n-4$. Note that not all edges are shown.}
\label{fig:relative_position_encodings}
\end{center}
\end{figure}

For linear sequences, edges can capture information about the relative position differences between input elements.
The maximum relative position we consider is clipped to a maximum absolute value of $k$.
We hypothesized that precise relative position information is not useful beyond a certain distance.
Clipping the maximum distance also enables the model to generalize to sequence lengths not seen during training.
Therefore, we consider $2k+1$ unique edge labels.

\begin{align*}
a^K_{ij} &= w^K_{\mathrm{clip}(j - i, k)} \\
a^V_{ij} &= w^V_{\mathrm{clip}(j - i, k)} \\
\mathrm{clip}(x, k) &= \max(-k, \min(k, x))
\end{align*}

We then learn relative position representations $w^K = (w^K_{-k}, \ldots, w^K_k)$ and
$w^V = (w^V_{-k}, \ldots, w^V_k)$ where $w^K_i, w^V_i \in \R^{d_a}$.

\subsection{Efficient Implementation}\label{secimpl}

There are practical space complexity concerns when considering
edges between input elements, as noted by Veli{\v{c}}kovi{\'c} et al.~\shortcite{velivckovic2017}, which considers unlabeled graph inputs to an attention model.

For a sequence of length $n$ and $\heads$ attention heads, we reduce the space complexity of storing relative position representations from $O(\heads n^2d_a)$ to $O(n^2d_a)$ by sharing them across each heads. Additionally, relative position representations can be shared across sequences. Therefore, the overall self-attention space complexity increases from $O(\batch \heads nd_z)$ to $O(\batch \heads nd_z + n^2d_a)$. Given $d_a = d_z$, the size of the relative increase depends on $\frac{n}{bh}$.

The Transformer computes self-attention efficiently for all sequences, heads, and positions in a batch using parallel matrix multiplication operations~\cite{vaswani2017}. 
Without relative position representations, each $e_{ij}$ can be computed using $bh$ parallel multiplications of $n \times d_z$ and $d_z \times n$ matrices. Each matrix multiplication computes $e_{ij}$ for all sequence positions, for a particular head and sequence. For any sequence and head, this requires sharing the same representation for each position across all compatibility function applications (dot products) with other positions.

When we consider relative positions the representations differ with different pairs of positions. This prevents us from computing all $e_{ij}$ for all pairs of positions in a single matrix multiplication. We also want to avoid broadcasting relative position representations. However, both issues can be resolved by splitting the computation of eq.~\eqref{eq:e2} into two terms:

\begin{equation}\label{eq:e3}
e_{ij} =  \frac{x_iW^Q(x_jW^K)^T + x_iW^Q(a^K_{ij})^T}{\sqrt{d_z}}
\end{equation}

The first term is identical to eq.~\eqref{eq:e}, and can be computed as described above.
For the second term involving relative position representations, tensor reshaping can be used to compute $n$ parallel multiplications of $bh \times d_z$ and $d_z \times n$ matrices. Each matrix multiplication computes contributions to $e_{ij}$ for all heads and batches, corresponding to a particular sequence position. Further reshaping allows adding the two terms. The same approach can be used to efficiently compute eq.~\eqref{eq:attn2}.

For our machine translation experiments, the result was a modest 7\% decrease in steps per second, but we were able to maintain the same model and batch sizes on P100 GPUs as Vaswani et al.~\shortcite{vaswani2017}.

\section{Experiments}

\begin{table*}[t!]

\begin{center}
\begin{tabular}{llll}
\hline
Model & Position Information & EN-DE BLEU & EN-FR BLEU \\
\hline
Transformer (base) & Absolute Position Representations & 26.5 & 38.2 \\
Transformer (base) & Relative Position Representations & \bf 26.8 & \bf 38.7 \\
\hline
Transformer (big) & Absolute Position Representations & 27.9 & 41.2 \\
Transformer (big) & Relative Position Representations & \bf 29.2 & \bf 41.5 \\
\hline
\end{tabular}
\end{center}
\caption{Experimental results for WMT 2014 English-to-German (EN-DE) and English-to-French (EN-FR) translation tasks, using newstest2014 test set.}
\label{mtresults}
\end{table*}

\subsection{Experimental Setup}

We use the tensor2tensor
\footnote{The tensor2tensor library is available at \url{https://github.com/tensorflow/tensor2tensor}.
}
library for training and evaluating our model.

We evaluated our model on the WMT 2014 machine translation task, using the WMT 2014 English-German dataset consisting of approximately 4.5M sentence pairs and
the 2014 WMT English-French dataset consisting of approximately 36M sentence pairs.

For all experiments, we split tokens into a 32,768 word-piece vocabulary ~\cite{wu2016}.
We batched sentence pairs by approximate length, and limited input and output tokens per batch to 4096 per GPU.
Each resulting training batch contained approximately 25,000 source and 25,000 target tokens.

We used the Adam optimizer ~\cite{kingma2014} with $\beta_1=0.9$, $\beta_2=0.98$, and $\epsilon = 10^{-9}$. We used the same warmup and decay strategy for learning rate as Vaswani et
al.~\shortcite{vaswani2017}, with 4,000 warmup steps.
During training, we employed label smoothing of value $\epsilon_{ls} = 0.1$ ~\cite{szegedy2016}.
For evaluation, we used beam search with a beam size of 4 and length penalty $\alpha = 0.6$ ~\cite{wu2016}.

For our base model, we used 6 encoder and decoder layers, $d_x = 512$, $d_z = 64$, 8 attention heads, 1024 feed forward inner-layer dimensions,
and $P_{dropout} = 0.1$.
When using relative position encodings, we used clipping distance $k = 16$,
and used unique edge representations per layer and head. We trained for 100,000 steps on 8 K40 GPUs,
and did not use checkpoint averaging.

For our big model, we used 6 encoder and decoder layers, $d_x = 1024$, $d_z = 64$, 16 attention heads, 4096 feed forward inner-layer dimensions, and $P_{dropout} = 0.3$ for EN-DE and $P_{dropout} = 0.1$ for EN-FR.
When using relative position encodings, we used $k = 8$, and used unique edge representations per layer.
We trained for 300,000 steps on 8 P100 GPUs, and averaged the last 20 checkpoints, saved at 10 minute intervals.

\subsection{Machine Translation}

We compared our model using only relative position representations to the baseline Transformer ~\cite{vaswani2017} with sinusoidal position encodings.
We generated baseline results to isolate the impact of relative position representations from any other changes to the underlying library and experimental configuration.

For English-to-German our approach improved performance over our baseline by 0.3 and 1.3 BLEU for the base and big configurations, respectively.
For English-to-French it improved by 0.5 and 0.3 BLEU for the base and big configurations, respectively.
In our experiments we did not observe any benefit from including sinusoidal position encodings in addition to relative position representations. The results are shown in Table \ref{mtresults}.

\subsection{Model Variations}\label{secmodelvariations}

We performed several experiments modifying various aspects of our model.
All of our experiments in this section use the base model configuration without any absolute position representations. BLEU scores are calculated on the WMT English-to-German task using the development set, newstest2013. 

We evaluated the effect of varying the clipping distance, $k$, of the
maximum absolute relative position difference. Notably, for $k \geq 2$, there does not appear to be much variation in BLEU scores. However, as we use multiple encoder layers, precise relative position information may be able to propagate beyond the clipping distance. The results are shown in Table \ref{clipping}.

\begin{table}[h]
\begin{center}
\scalebox{0.9}{
\begin{tabular}{|l|l|}
\hline
$k$ & EN-DE BLEU \\
\hline
0 & 12.5 \\
1 & 25.5 \\
2 & 25.8 \\
4 & 25.9 \\
16 & 25.8 \\
64 & 25.9 \\
256 & 25.8 \\
\hline
\end{tabular}
}
\end{center}
\caption{
Experimental results for varying the clipping distance, $k$.}
\label{clipping}
\end{table}

We also evaluated the impact of ablating each of the two relative position representations defined in section \ref{secselfattn}, $a^V_{ij}$ in eq.~\eqref{eq:attn2} and $a^K_{ij}$ in eq.~\eqref{eq:e2}. Including relative position representations solely when determining compatibility between elements may be sufficient, but further work is needed to determine whether this is true for other tasks. The results are shown in Table \ref{edges}.

\begin{table}[h]
\begin{center}
\scalebox{0.9}{
\begin{tabular}{|l|l|l|}
\hline
 & & \\[-1em] 
$a^V_{ij}$  & $a^K_{ij}$ & EN-DE BLEU \\

\hline
Yes & Yes & 25.8 \\
No & Yes & 25.8 \\
Yes & No & 25.3 \\
No & No & 12.5 \\
\hline
\end{tabular}
}
\end{center}
\caption{
\label{edges}
Experimental results for ablating relative position representations $a^V_{ij}$ and $a^K_{ij}$.
}
\end{table}

\section{Conclusions}

In this paper we presented an extension to self-attention that can be used to incorporate relative position information for sequences, which improves performance for machine translation.

For future work, we plan to extend this mechanism to consider arbitrary directed, labeled graph inputs to the Transformer.
We are also interested in nonlinear compatibility functions to combine input representations and edge representations.
For both of these extensions, a key consideration will be determining efficient implementations.

\bibliographystyle{acl_natbib}
\bibliography{paper}

\end{document}